\documentclass[jimaging,article,accept,pdftex,moreauthors]{Definitions/mdpi} 

\firstpage{1} 
\pubvolume{1}
\issuenum{1}
\articlenumber{0}
\pubyear{2023}
\copyrightyear{2023}
\datereceived{5 May 2023  } 
\daterevised{22 May 2023 } 
\dateaccepted{25 May 2023} 
\datepublished{ } 
\hreflink{https://doi.org/} 
\usepackage{comment}

\Title{Gender, Smoking History and Age Prediction from Laryngeal~Images}

\TitleCitation{Gender, Smoking History and Age Prediction from Laryngeal~Images}

\Author{ {Tianxiao Zhang }$^{1}$\orcidA, Andrés  {M.} Bur $^{2}$, Shannon Kraft $^{2}$, Hannah Kavookjian $^{2}$, Bryan Renslo $^{2}$, Xiangyu Chen $^{1}$, Bo~Luo $^{1}$ and Guanghui Wang $^{3,}$*}

\AuthorNames{Tianxiao Zhang, Andrés M. Bur,  Shannon Kraft, Hannah Kavookjian, Bryan Renslo, Xiangyu Chen, Bo~Luo and Guanghui Wang}

\AuthorCitation{Zhang, T.; M. Bur, A.;  Kraft, S.; Kavookjian, H.; Renslo, B.; Chen, X.; Luo, B.; Wang, G.}

\address{%
$^{1}$ \quad Department of Electrical Engineering and Computer Science, University of Kansas, Lawrence,  {KS,} 
 USA;  {tianxiao@ku.edu (T.Z.); xychen@ku.edu (X.C.); bluo@ku.edu (B.L.)}\\ 
$^{2}$ \quad Department of Otolaryngology  {-} Head and Neck Surgery, University of Kansas Medical Center, \mbox{Kansas City,  {KS,} 
 USA};  {abur@kumc.edu (A.M.B.); skraft3@kumc.edu (S.K.); hkavookjian@kumc.edu (H.K.); brenslo@kumc.edu (B.R.)}\\
$^{3}$ \quad Department of Computer Science, Toronto Metropolitan University, Toronto,  {ON,} 
 Canada}

\corres{Correspondence: wangcs@torontomu.ca}

\abstract{Flexible laryngoscopy is commonly performed by otolaryngologists to detect laryngeal diseases and to recognize potentially malignant lesions. Recently, researchers have introduced machine learning techniques to facilitate automated diagnosis using laryngeal images and achieved promising results. Diagnostic performance can be improved when patients' demographic information is incorporated into models. However, manual entry of patient data is time consuming for clinicians. In this study, we made the first endeavor to employ deep learning models to predict patient demographic information to improve detector model performance. The overall accuracy for gender, smoking history, and age was 85.5\%, 65.2\%, and 75.9\%, respectively. We also created a new laryngoscopic image set for machine learning study and benchmarked the performance of 8 classical deep learning models based on CNNs and Transformers. The results can be integrated into current learning models to improve their performance by incorporating the patient's demographic information.}

\keyword{Laryngeal images; CAM; gender; smoking history; demographic information} 

\begin{document}


\section{Introduction}

Flexible laryngoscopy is a commonly used diagnostic tool to visually identify diseases of the larynx~\cite{leipzig_role_1985,ebisumoto_tumor_2021}. While it has advantages over other diagnostic methods given its ease of use and lack of ionizing radiation exposure, discerning between benign and malignant lesions on laryngoscopy requires expert interpretation. Previously, computer vision techniques utilizing deep learning, including Convolutional Neural Networks (CNNs) or Transformers, have been implemented to determine pathologic diagnosis based on laryngoscopic medical images or video~\cite{xiong2019computer,halicek2017deep,azam2022deep,takiyama2018automatic,wang2022hierarchical}. Such models have shown to be sufficiently accurate in the diagnosis of laryngeal cancer with only a limited training set~\cite{xiong2019computer,halicek2017deep,azam2022deep,takiyama2018automatic,wang2022hierarchical,ren2020automatic}.

The majority of prior studies that utilize machine learning for medical image analysis focus on lesion or polyp detection, segmentation, and classification~\cite{wilson2022harnessing,li2021colonoscopy,patel22fuzzynet,patel2021enhanced,patel2020comparative}. To date, no studies have attempted to automatically incorporate patient characteristics into lesion detection models by predicting them using  laryngeal images. Even for well-trained experts, identifying the age, gender, or smoking status of patients based on laryngoscopy alone is virtually impossible. Fortunately, this is never necessary because this information is readily available to clinicians performing laryngoscopy. However, incorporation of patient characteristics into deep learning models for medical image analysis typically requires manual entry. 

In this study, we have demonstrated the capability of deep learning models, such as CNNs and Transformers, to extract discernible features from laryngeal images, allowing the identification of patients' demographic characteristics. This has the potential to enhance clinical diagnosis by automatically integrating demographic information into intelligent learning models. For instance, we can automate multi-model learning to improve the detection of laryngeal cancers by considering factors like the patient's smoking status and age during decision-making. Additionally, our research contributes to the field of explainable machine learning (XAI), which emphasizes the provision of clear and interpretable explanations for the decisions and predictions of models. By enhancing transparency and trust, XAI plays a crucial role in medical contexts where healthcare decisions carry significant importance~\cite{militello2023ct,gu2020net,van2022explainable,prinzi2023ml}. Analyzing patients' demographic characteristics, especially the activation saliency maps, can deepen our understanding of the underlying workings of deep learning models.

This study makes the first endeavor to predict the patient's gender, smoking history, and age directly from laryngeal images. We have implemented and compared the performance of the following classical CNN-based and Transformer-based deep learning models: ResNet-18~\cite{he2016deep}, ResNet-50~\cite{he2016deep}, ResNet-101~\cite{he2016deep}, DenseNet-121~\cite{huang2017densely}, MobileNetv2~\cite{sandler2018mobilenetv2}, ShuffleNetv2~\cite{ma2018shufflenet}, and ViT~\cite{dosovitskiy2020image}. The major contributions of this paper are as below:
\begin{itemize}
    \item We performed the first study on predicting  the gender, age, and smoking status of the patient purely based on laryngeal images from laryngoscopy.
    \item We created a dataset of 33,906 laryngeal image frames captured from 398 patients. The dataset is annotated with clinical diagnosis, pathologic diagnosis for lesion, and patient demographic information. This is the first large laryngoscopic image set for machine learning studies.
    \item We implemented and benchmarked the performance of 8 classical deep learning models and achieved very promising results. 
    \item We employ the Classification Activation Map (CAM) to visualize and analyze the regions of interest in the image. This approach contributes to the explainability of the learning models by providing insights into which specific areas of the image influenced the decision-making process. 
\end{itemize}

The labeled dataset and developed learning models are available to the research community upon request.

\section{Materials and Methods}

\subsection{Dataset}

Data from flexible video stroboscopic exams performed during patient care in the Department of Otolaryngology-Head \& Neck Surgery at the University of Kansas Medical Center (KUMC) were collected over a one-year period. Digital videos were collected in MPEG-4 format at 30 frames per second (fps) with a resolution of 720 $\times$ 486 pixels. Each video was labeled with a clinical diagnosis (structurally normal larynx, polyp, papilloma, leukoplakia, or malignant neoplasm) and a pathologic diagnosis for lesions that were biopsied. Additional patient demographic information was captured including age, sex, and history of tobacco use.

A total of 398 video sequences were included for analysis and randomly separated into training (n = 319, 80\%) and testing (n = 79, 20\%) cohorts. Every 10th video frame was extracted from each video sequence, creating a dataset of 33,906 laryngeal images in total with 26,424 for training and 7,482 for testing. All classification models were pretrained on the ImageNet benchmark~\cite{russakovsky2015imagenet}. Transfer learning was then used to fine-tune the learning models using the collected training set. Finally, the testing set was employed to evaluate the performance of the final classification models.



\subsection{Deep Learning Models}

\begin{figure}[H]
\centering
\includegraphics[width=1\linewidth]{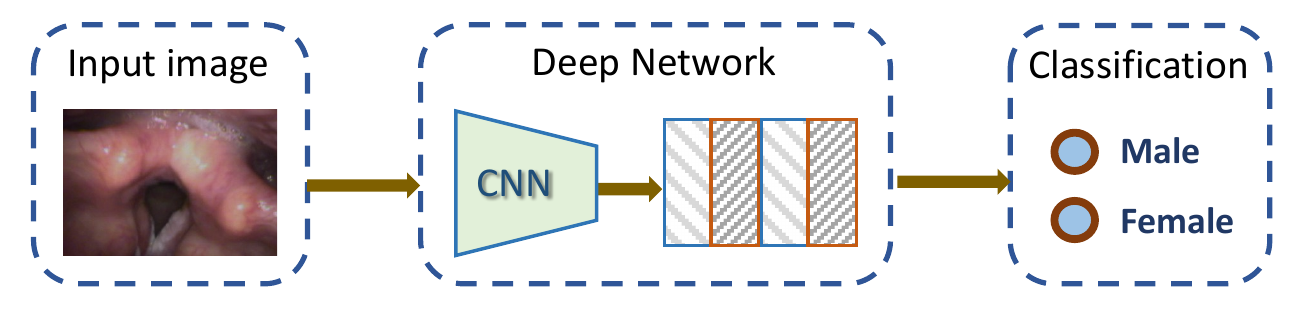}
\caption{Illustration of using deep learning models for laryngeal image classification. The deep learning models are pre-trained on ImageNet and then fine-tuned on the laryngeal dataset using transfer learning. The output prediction could be gender, smoking history, or age.}
\label{fig:0}
\end{figure}

The following classical deep learning models were implemented and compared: ResNet-18~\cite{he2016deep}, ResNet-50~\cite{he2016deep}, ResNet-101~\cite{he2016deep}, DenseNet-121~\cite{huang2017densely}, MobileNetv2~\cite{sandler2018mobilenetv2}, ShuffleNetv2~\cite{ma2018shufflenet}, and Vision Transformer~\cite{dosovitskiy2020image}. The general process of deep learning based classification is shown in Figure~\ref{fig:0}. Given an input of a laryngeal image frame, the trained network can predict the gender, smoking history, or age of the patient purely based on the features in the input image. Below is a brief introduction the implemented learning models.

\noindent\textbf{ {ResNet:}} ResNet~\cite{he2016deep} designs a residual connection to facilitate the training of deep neural networks. The gradients could be easier back-propagated via the short connections so that the deep neural networks could be optimized more easily and have better performance than their shallow counterparts. Since its introduction, ResNet has become the benchmark for almost all computer vision tasks and has achieved state-of-the-art performance in almost all tasks. Additionally, shortcut connections can be applied to other classic models such as Transformers to achieve state-of-the-art performance in both natural language processing and computer vision applications.

\noindent {\textbf{DenseNet:}} In DenseNet~\cite{huang2017densely}, the layers are connected with each other directly so that the gradient can flow smoothly, preventing information flow from vanishing, which is a common difficulty in deep neural network training. The features from different layers are combined by concatenation instead of summation.

\noindent {\textbf{MobileNetv2:}} MobileNetv2~\cite{sandler2018mobilenetv2} is based on MobileNetv1~\cite{howard2017mobilenets} which separates the convolutions into depthwise separable convolutions and pointwise convolutions with fewer parameters and computations. MobileNetv2 introduces an inverted residual block that projects the feature maps to a high dimension and then back to a low dimension. The proposed inverted module reduces memory access and accelerates inference speed.

\noindent {\textbf{ShuffleNetv2:}} ShuffleNetv2~\cite{ma2018shufflenet} was developed from ShuffleNetv1~\cite{zhang2018shufflenet} to empirically design high-efficient mobile-level networks. Practical guidelines were incorporated for higher efficiency and a more lightweight network, including equal channel width, group convolution cost, less network fragmentation, and fewer element-wise operations.

\noindent {\textbf{Vision Transformers:}} Transformers~\cite{vaswani2017attention} were initially designed for natural language processing for global connections between long-range tokens. Transformers have since been applied to computer vision tasks and have achieved state-of-the-art performance in classification~\cite{chen2023accumulated,liu2021swin} and object detection~\cite{carion2020end,ma2021miti}. For image classification, the images are split into patches of the same size, which are embedded into tokens and fed into the Transformer blocks. Usually, there is an extra class token that interacts with all other tokens and produces the ultimate class prediction. Due to the lack of inductive bias, vision Transformers~\cite{dosovitskiy2020image} normally require more data and much longer training epochs to converge.

\subsection{Training Settings}

Given that the current dataset was relatively small compared to other benchmark datasets in computer vision, transfer learning was employed and all deep learning models were pre-trained on ImageNet~\cite{russakovsky2015imagenet}. For each learning model, the same structure and hyperparameters as reported in the original paper were utilized. The batch size was set to 16 and the initial learning rate was 0.00005 (reduced by 0.2 each epoch) with a total of 5~epochs. The optimizer utilized was Adam~\cite{kingma2014adam}, and all code was written with PyTorch~\cite{paszke2019pytorch}. 

\subsection{The Metrics for Evaluation}
We evaluate the performance of our deep learning models on the laryngeal dataset using four commonly used metrics: Precision, recall, F1 score, and overall accuracy. The definitions of these metrics can be found in~\cite{patel2020comparative}.

Precision assesses the accuracy of positive predictions by measuring the proportion of correctly classified positive instances out of all instances predicted as positive. It indicates how well the model identifies positive instances and has a low false positive rate. Recall, also known as sensitivity or true positive rate, measures the proportion of correctly classified positive instances out of all actual positive instances. It focuses on the model's ability to detect all positive instances and has a low false negative rate. The F1 score combines precision and recall into a single value, providing a balanced measure. It is calculated as the harmonic mean of precision and recall, considering both false positives and false negatives. The F1 score serves as an overall performance metric, providing a single evaluation measure. Overall accuracy measures the proportion of correctly classified instances, including both positive and negative, out of all instances.

These metrics offer insights into different aspects of the model's classification abilities. When evaluating a medical image classification model, it is crucial to consider the specific requirements and priorities of the application. The importance of each metric may vary depending on the context. Additionally, it is important to interpret these metrics alongside domain-specific considerations, such as the severity of misclassifications and their potential impact on patient outcomes.

\section{Results}

A total of 398 video sequences were utilized in our analysis, which were further divided into two cohorts: a training cohort consisting of 319 sequences and a testing cohort comprising 79 sequences. The models were trained using the training cohort, taking into account the ground truth information regarding the patient's gender, smoking history, and age. Subsequently, the classification performance of the models was assessed using the independent testing set. In this section, we begin by evaluating the model's performance at the image level and subsequently present the results at the patient (sequence) level. This approach allows us to examine both the individual image classification accuracy and the overall performance across the entire video sequence.

\subsection{The Performance of Deep Learning Models at Image Level}

\begin{figure}[H]

\begin{adjustwidth}{-\extralength}{0cm}\centering
\includegraphics[width=.438\textwidth]{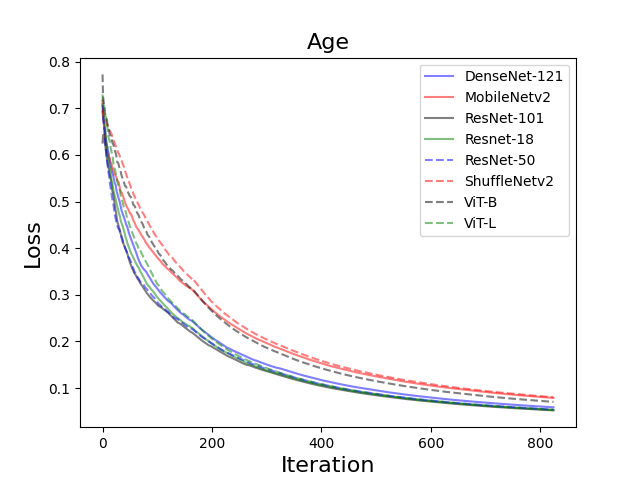}
\includegraphics[width=.438\textwidth]{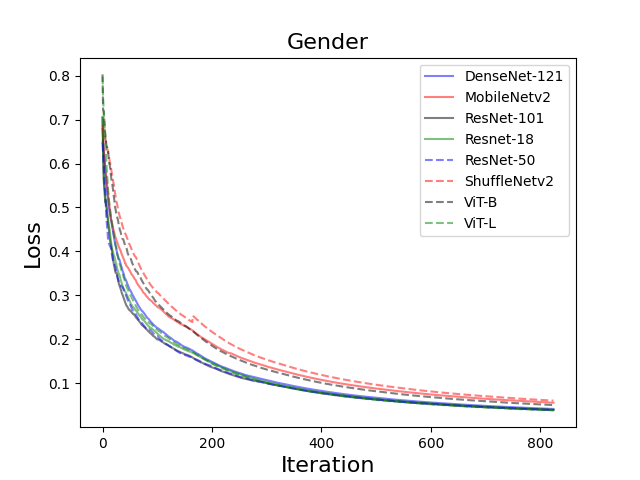}
\includegraphics[width=.438\textwidth]{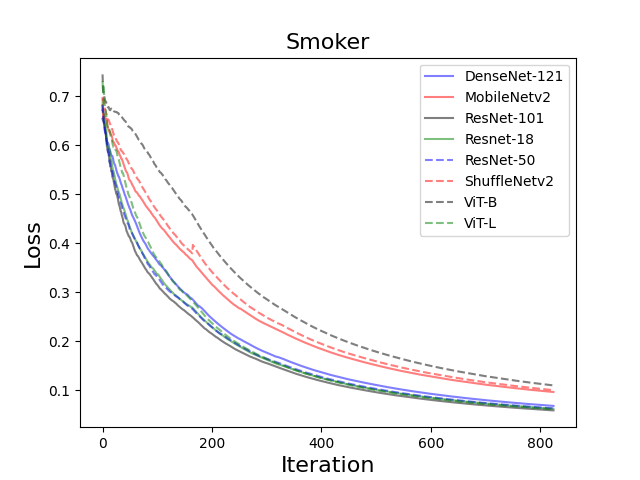}
\end{adjustwidth}

\caption{ {The average} loss curve for training the models to predict age, gender, and smoke history. The left, middle, and right graphs are the loss curve for predicting age, gender, and smoke history,~respectively.}
\label{fig:-1} 
\end{figure}

Figure~\ref{fig:-1} depicts the loss curves of different models employed for age, gender, and smoking history prediction. The loss measure utilized in the analysis is the average loss calculated across all previous loss values, resulting in a smoothed loss curve. During the training process, the loss curves converge quickly, and the training is terminated after five epochs. It is observed that MobileNetv2, ShuffleNetv2, and ViT-B exhibit relatively higher loss values compared to the other models at convergence.

The evaluation metrics, including precision, recall, F1 score, and overall accuracy, are presented in Tables~\ref{table:1}--\ref{table:4}. These metrics assess the performance of the models in predicting three target categories: gender (male or female), smoking history (smoker or non-smoker), and age ($<50$ or $\ge50$). Each category represents a binary classification problem. In the case of smoking history, a non-smoker is defined as a patient who has never smoked, while a smoker refers to a patient with any smoking history. For age prediction, patients are divided into two groups: young ($<50$) and senior ($\ge50$), creating a binary classification scenario. The reported precision, recall, F1 score, and overall accuracy provide a comprehensive assessment of the models' performance across these classification tasks.


\begin{table}[H]
\centering
\caption{ {The precision} of predicting gender, smoking history, and age on the larynx dataset}
\begin{tabularx}{\textwidth}{cCCCcCC}
 \toprule
 \multirow{2}{*}{\vspace{-4pt}\textbf{DL Models}} & \multicolumn{2}{c}{\textbf{Gender}} & 
     \multicolumn{2}{c}{\textbf{Smoke History}} & \multicolumn{2}{c}{\textbf{Age}} \\
 \cmidrule{2-7}
 & \textbf{Male} & \textbf{Female} & \textbf{Smoker} & \textbf{Non-Smoker} & \boldmath{$<50$ }& \boldmath{$\ge50$} \\ 
 \midrule
ResNet-18 & 93.9 & 73.4 & 65.3 & 63.3 & 41.6 & 88.6 \\

ResNet-50 & 94.7 & 70.9 & 64.2 & 62.8 & 44.5 & 87.8 \\

ResNet-101 & 92.4 & 70.9 & 66.7 & 64.0 & 45.7 & 89.7\\

DenseNet-121 & 94.6 & 72.3 & 62.0 & 60.0 & 44.3 & 89.5\\

MobileNetv2 & 93.6 & 66.1 & 64.1 & 61.5 & 43.3 & 89.5\\

ShuffleNetv2 & 92.0 & 66.8 & 64.6 & 63.1 & 38.5 & 88.0 \\

ViT-L & 93.5 & 66.4 & 66.6 & 63.1 & 47.7 & 89.1 \\

ViT-B & 94.3 & 69.4 & 64.7 & 63.6 & 46.6 & 88.4 \\
\midrule

Mean & 93.6 & 69.5 & 64.8 & 62.7 & 44.0 & 88.8 \\

Std & 0.99 & 2.81 & 1.50 & 1.30 & 2.93 & 0.73 \\

\bottomrule
 \end{tabularx}
 \label{table:1}
 \end{table}

Table~\ref{table:1} provides the precision values calculated for each class, along with the mean and standard deviation computed across all deep learning models. Overall, the models exhibited consistent performance across the experiments, although the standard deviations for predicting female gender and age $<50$ were relatively large. Among all deep learning models, ResNet-50 achieved the highest precision for predicting male gender, with a value of 94.7\%. In contrast, the precision for age $<50$ was significantly lower at only 44\%, compared to the other categories. The high precision for predicting male gender can be attributed to the distinguishable features between male and female patients, as well as the clear visual differences between male and female images. In contrast, discerning features related to age becomes more challenging, particularly for patients near the age threshold.

\begin{table}[H]
\caption{ {The recall} of predicting gender, smoking history, and age on the larynx dataset}
\begin{tabularx}{\textwidth}{cCCCcCC}
 \toprule
 \multirow{2}{*}{\vspace{-4pt}\textbf{DL Models}} & \multicolumn{2}{c}{\textbf{Gender}} & 
     \multicolumn{2}{c}{\textbf{Smoke History}} & \multicolumn{2}{c}{\textbf{Age}} \\
 \cmidrule{2-7}
 & \textbf{Male} & \textbf{Female} & \textbf{Smoker} & \textbf{Non-Smoker} & \boldmath{$<50$ }& \boldmath{$\ge50$} \\ 
 \midrule
ResNet-18 & 83.5 & 89.3 & 59.6 & 68.7 & 68.7 & 71.7 \\

ResNet-50 & 81.1 & 91.0 & 59.8 & 67.1 & 63.6 & 76.7 \\

ResNet-101 & 81.9 & 86.8 & 59.8 & 70.6 & 70.6 & 75.4\\

DenseNet-121 & 82.3 & 90.8 & 54.8 & 66.8 & 70.5 & 74.0\\

MobileNetv2 & 76.7 & 89.6 & 56.3 & 68.9 & 71.1 & 72.7\\

ShuffleNetv2 & 78.1 & 86.6 & 59.9 & 67.6 & 68.4 & 67.9 \\

ViT-L & 76.9 & 89.5 & 57.7 & 71.4 & 67.3 & 78.4 \\

ViT-B & 79.8 & 90.5 & 61.2 & 67.0 & 65.0 & 78.2 \\
\midrule

Mean & 80.0 & 89.3 & 58.6 & 68.5 & 68.2 & 74.4 \\

Std & 2.58 & 1.70 & 2.17 & 1.73 & 2.72 & 3.57 \\

\bottomrule
 \end{tabularx}
 \label{table:2}
 \end{table}
 
 \vspace{-6pt}
\begin{table}[H]
\caption{The F1 score of predicting gender, smoking history, and ages on the larynx dataset}
\begin{tabularx}{\textwidth}{cCCCcCC}
 \toprule
\multirow{2}{*}{\vspace{-4pt}\textbf{DL Models}} & \multicolumn{2}{c}{\textbf{Gender}} & 
     \multicolumn{2}{c}{\textbf{Smoke History}} & \multicolumn{2}{c}{\textbf{Age}} \\
 \cmidrule{2-7}
 & \textbf{Male} & \textbf{Female} & \textbf{Smoker} & \textbf{Non-Smoker} & \boldmath{$<50$ }& \boldmath{$\ge50$} \\ 
 \midrule
ResNet-18 & 88.4 & 80.6 & 62.3 & 65.9 & 51.8 & 79.3 \\

ResNet-50 & 87.3 & 79.7 & 61.9 & 64.9 & 52.4 & 81.9 \\

ResNet-101 & 86.9 & 78.1 & 63.1 & 67.1 & 55.5 & 82.0\\

DenseNet-121 & 88.0 & 80.5 & 58.2 & 63.2 & 54.4 & 81.0\\

MobileNetv2 & 84.3 & 76.1 & 60.0 & 65.0 & 53.8 & 80.2\\

ShuffleNetv2 & 84.5 & 75.4 & 62.2 & 65.2 & 49.3 & 76.7 \\

ViT-L & 84.4 & 76.2 & 61.8 & 67.0 & 55.9 & 83.4 \\

ViT-B & 86.4 & 78.6 & 62.9 & 65.3 & 54.3 & 83.0 \\
\midrule

Mean & 86.3 & 78.2 & 61.6 & 65.5 & 53.4 & 80.9 \\

Std & 1.67 & 2.06 & 1.65 & 1.25 & 2.17 & 2.19 \\

\bottomrule
 \end{tabularx}
 \label{table:3}
 \end{table}
 
  \vspace{-6pt}
 
\begin{table}[H]
\centering
\caption{The overall accuracy of predicting gender, smoking history, and age on the larynx dataset}
\begin{tabularx}{\textwidth}{CCCC}
 \toprule
\textbf{DL Models} & \textbf{Gender }& \textbf{Smoking History} & \textbf{Age} \\
 \midrule
ResNet-18 & 85.5 & 64.2 & 71.0 \\

ResNet-50 & 84.4 & 63.5 & 73.7 \\

ResNet-101 & 83.6 & 65.2 & 74.3 \\

DenseNet-121 & 85.2 & 60.9 & 73.2 \\

MobileNetv2 & 81.0 & 62.6 & 72.3 \\

ShuffleNetv2 & 81.0 & 63.8 & 68.0 \\

ViT-L & 81.2 & 64.6 & 75.9 \\

ViT-B & 83.4 & 64.1 & 75.2 \\
\midrule

Mean & 83.2 & 63.6 & 73.0 \\

Std & 1.87 & 1.34 & 2.53 \\

\bottomrule
 \end{tabularx}
 \label{table:4}
 \end{table}

The standard deviation of accuracy among the models, as shown in Table~\ref{table:1}, indicates that there is relatively low variation in performance across the different models. Notably, the lightweight deep learning models, such as ResNet-18, outperform the more complex models with a larger number of parameters (e.g., ResNet-18 achieving the highest precision for predicting ``female''). This observation suggests that the limited size of the dataset may favor simpler models, as they are less prone to overfitting. The dataset's relatively small size may also contributComplex models like Vision Transformers typically require a larger amount of data to achieve optimal performance. While precision provides valuable insights into the models' performance, it is important to consider other metrics as well. The following sections will present the models' performance based on additional evaluation~metrics.

Table~\ref{table:2} provides the recall rates for each class. The recall rate measures the proportion of positive samples that are correctly identified among all positive samples in each category. While the recall rates exhibit relatively higher variations compared to precision, the results remain consistent across all models. Notably, smoking history exhibits the lowest recall rate among all deep learning models. This can be attributed, in part, to the inherent variability in smoking habits among individuals. Some smokers who have minimal smoking frequency or have quit smoking for an extended period may display fewer visible changes in their larynx, making it challenging to distinguish them from non-smokers solely based on visual cues. As a result, accurately identifying these individuals as smokers becomes more difficult, leading to a lower recall rate for smoking history prediction.


To comprehensively evaluate the performance of the deep learning models on the larynx dataset, it is important to consider metrics that incorporate both precision and recall. The F1 score, as presented in Table~\ref{table:3}, computes the harmonic mean of precision and recall, providing a balanced assessment of the models. The F1 scores among the different deep learning models exhibit consistency, as indicated by the small standard deviations. Notably, the performance for predicting male gender, female gender, and age $\geq 50$ surpasses that of other classes in terms of F1 score. This implies that the models achieve a good balance between precision and recall for these categories, resulting in higher overall performance.


The overall accuracy of each learning model was assessed by calculating the number of correctly predicted samples divided by the total number of samples. Table~\ref{table:4} presents the results, showing that gender prediction achieved the highest overall accuracy, followed by age and smoking history predictions. Notably, gender prediction exhibited a particularly high mean accuracy among the three tasks, with an average overall predicted accuracy of 83.2\%. The impressive accuracy in gender prediction suggests that deep learning models can effectively capture and analyze specific features present in laryngeal images that are indicative of a patient's gender. These distinguishing features may not be readily discernible to human experts, underscoring the potential of deep learning models in extracting valuable information from medical images.


Gender prediction presents a straightforward binary classification task, whereas age and smoking history classification pose more significant challenges due to their continuous nature. Dividing age into specific thresholds becomes difficult as the distinguishing features between different age groups may not be readily apparent. Similarly, predicting smoking history is complex due to the wide range of addiction levels among smokers. For instance, the characteristics of a social smoker or someone with a short smoking history may differ significantly from those of a heavy smoker. Consequently, the boundaries between smokers and non-smokers are not always clearly discernible, despite the existence of distinct boundaries between heavy smokers and non-smokers. These challenges in establishing clear boundaries likely contribute to the relatively lower accuracy observed in age prediction compared to gender prediction.

Despite these inherent difficulties, the developed learning models still achieved notable mean accuracies of 73\% for age classification and 63.6\% for smoking history prediction. These results demonstrate the models' capability to capture meaningful patterns and extract relevant information from the laryngeal images, enabling reasonably accurate predictions. Although classifying age and smoking history entails inherent complexities, the achieved accuracies indicate that the models have successfully learned and utilized discriminative features to make informed predictions in these challenging tasks. These findings highlight the potential of machine learning in extracting valuable information from laryngeal images for age and smoking history classification.

In summary, deep learning models demonstrate strong performance in predicting gender, smoking history, and age, with gender prediction being particularly notable. The models surpass human doctors in extracting this information solely from laryngeal images, showcasing their potential in advancing medical image analysis. This finding underscores the promising role of deep learning models in leveraging visual data to enhance diagnostic capabilities in healthcare. By effectively identifying subtle patterns and characteristics, these models can aid healthcare professionals in providing more accurate assessments based on laryngeal images, ultimately improving patient care and outcomes.

\subsection{Overall Performance Based On Patients}

The experiments conducted above are evaluated based on individual image frames, but in clinical settings, all frames in a video sequence belong to the same patient. Therefore, it is more meaningful to evaluate the performance of classification at the sequence level. This section reports the overall accuracy of gender, smoking history, and age prediction at the patient level by combining the results of all frames in the same sequence.

Two methods are used to evaluate sequence-level prediction: majority voting and probability voting. In majority voting, the final prediction is based on the majority of the predicted image labels in the sequence. In probability voting, the predicted probabilities for the correct and wrong labels are separately aggregated, and the final prediction is assigned to the one with higher aggregated probabilities. The comparative results are presented in Table~\ref{table:5}. It is evident that using sequence-based prediction, the overall accuracy for predicting gender, smoking history, and age is much higher than that based on individual frames shown in Table~\ref{table:4}. We also notice that the overall performance of majority-based voting is slightly better than that of the probability-based approach.

 \begin{table}[H]
\caption{The overall accuracy for patients}
\begin{tabularx}{\textwidth}{cCCCCCC}
 \toprule
 \multirow{2}{*}{\vspace{-4pt}\textbf{DL Models}} & \multicolumn{2}{c}{\textbf{Gender}} & 
     \multicolumn{2}{c}{\textbf{Smoking History}} & \multicolumn{2}{c}{\textbf{Age}} \\
 \cmidrule{2-7} 
& \textbf{Majority} & \textbf{Prob} & \textbf{Majority} & \textbf{Prob} & \textbf{Majority} & \textbf{Prob}  \\
 \midrule
ResNet-18 & 90.7 & 88.9 & 66.9 & 62.5 & 77.5 & 73.1 \\

ResNet-50 & 88.9 & 86.5 & 62.1 & 59.1 & 81.0 & 78.2 \\

ResNet-101 & 84.5 & 84.7 & 64.8 & 62.3 & 77.8 & 74.4 \\

DenseNet-121 & 88.0 & 87.0 & 61.7 & 58.3 & 83.6 & 79.0 \\

MobileNetv2 & 84.5 & 81.2 & 63.2 & 59.7 & 77.5 & 73.5 \\

ShuffleNetv2 & 83.7 & 79.8 & 67.5 & 63.0 & 73.5 & 70.4 \\

ViT-L & 84.3 & 82.2 & 69.1 & 66.6 & 83.5 & 80.6 \\

ViT-B & 91.1 & 89.1 & 68.3 & 66.8 & 85.7 & 82.3 \\
\midrule

Mean & 87.0 & 84.9 & 65.4 & 62.3 & 80.0 & 76.5 \\

Std & 2.87 & 3.31 & 2.71 & 3.00 & 3.85 & 3.89 \\

\bottomrule
 \end{tabularx}
 \label{table:5}
 \end{table}

\subsection{Visualization}

\begin{figure}[H]
\includegraphics[width=.48\textwidth]{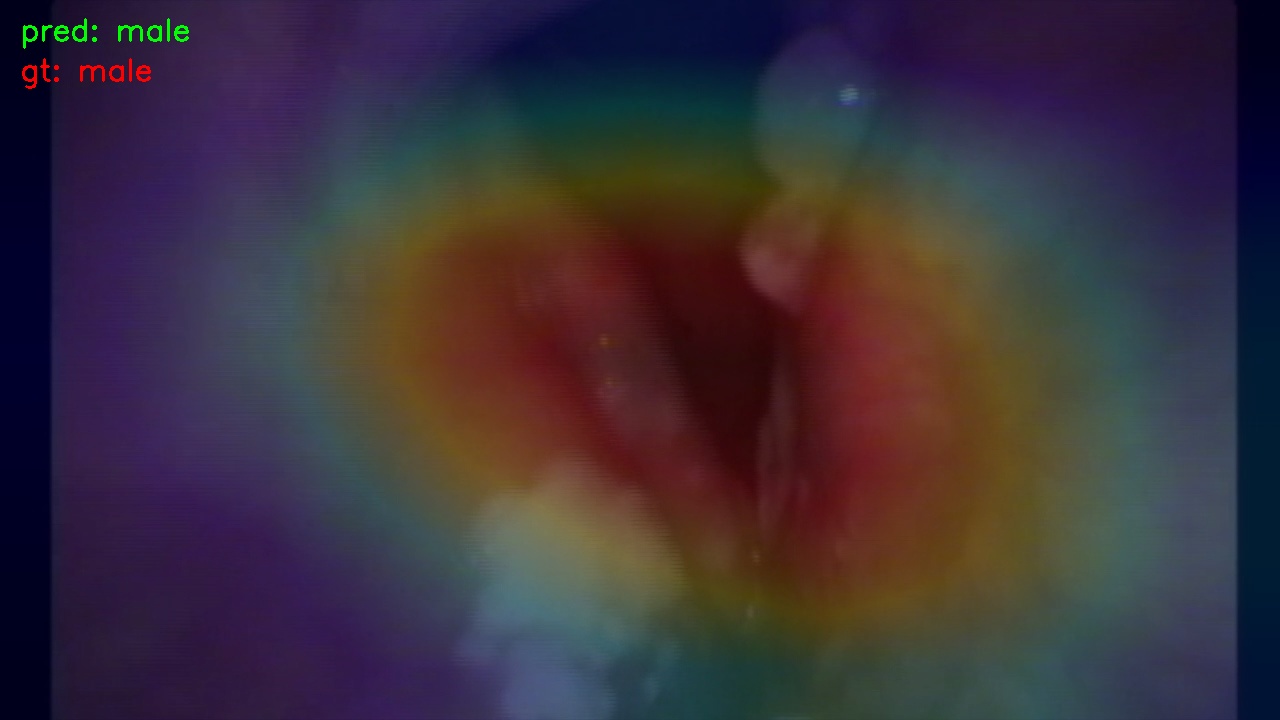}
\includegraphics[width=.48\textwidth]{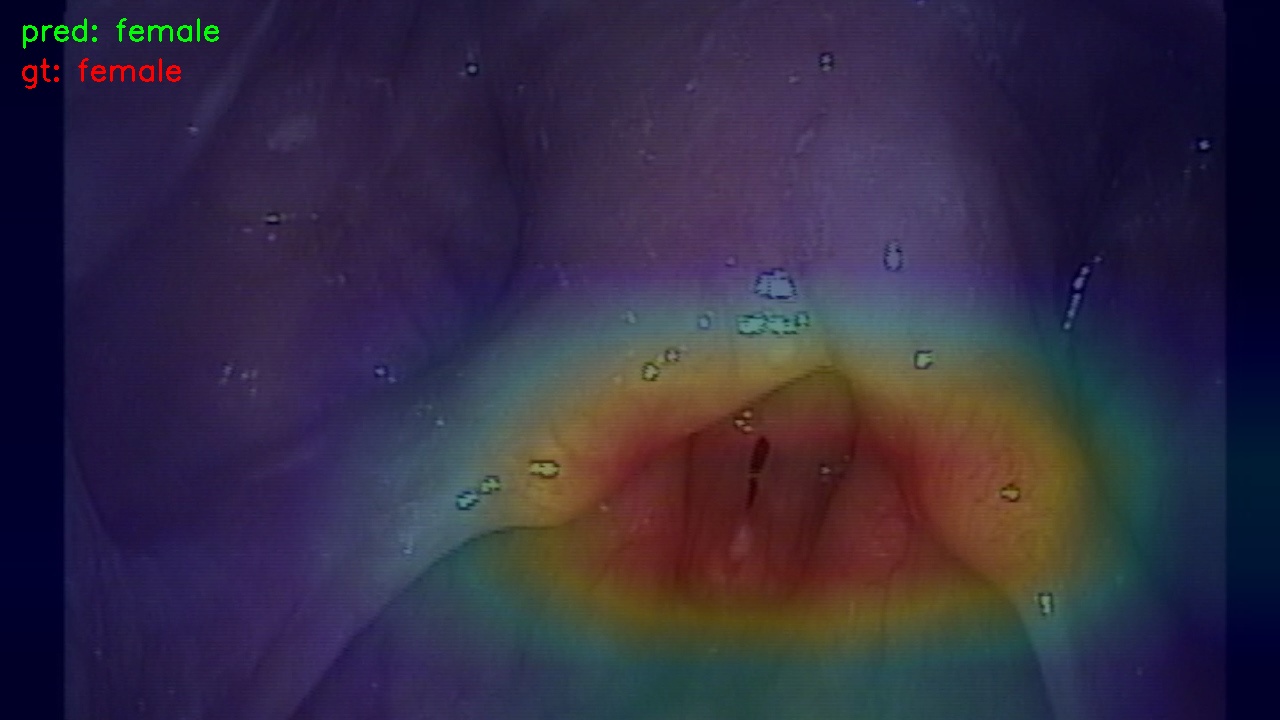}\\
\includegraphics[width=.48\textwidth]{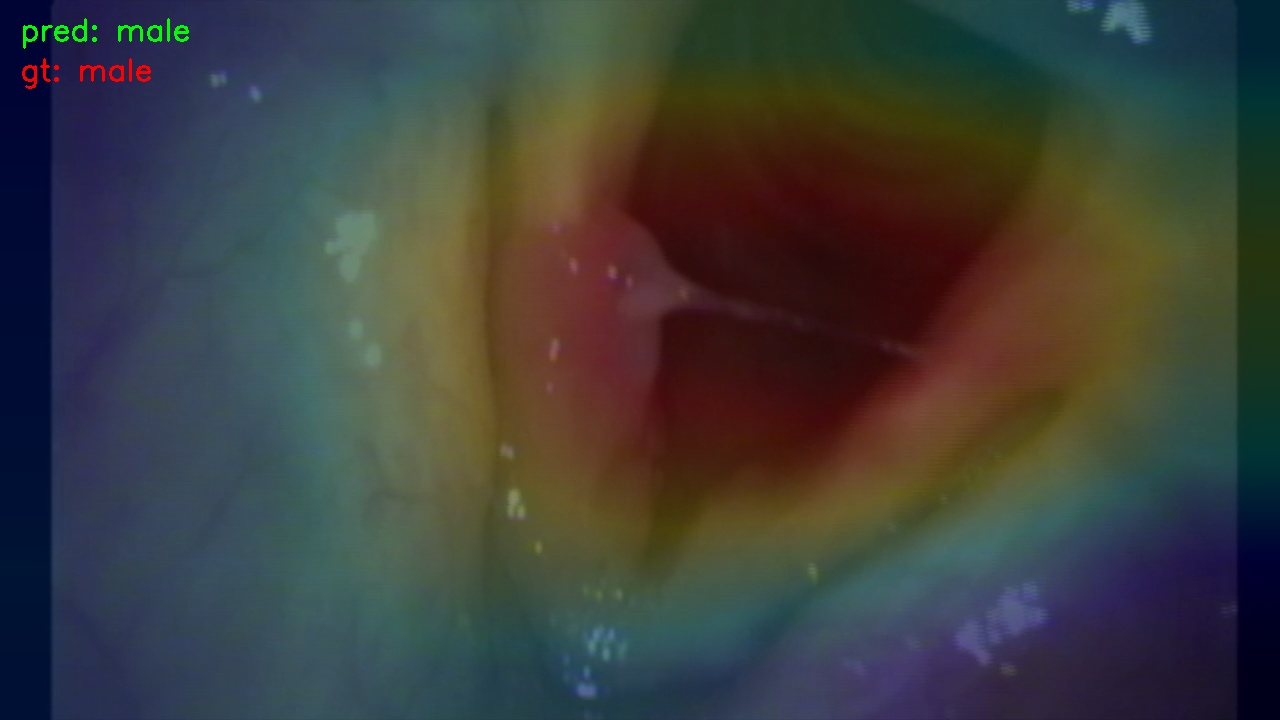}
\includegraphics[width=.48\textwidth]{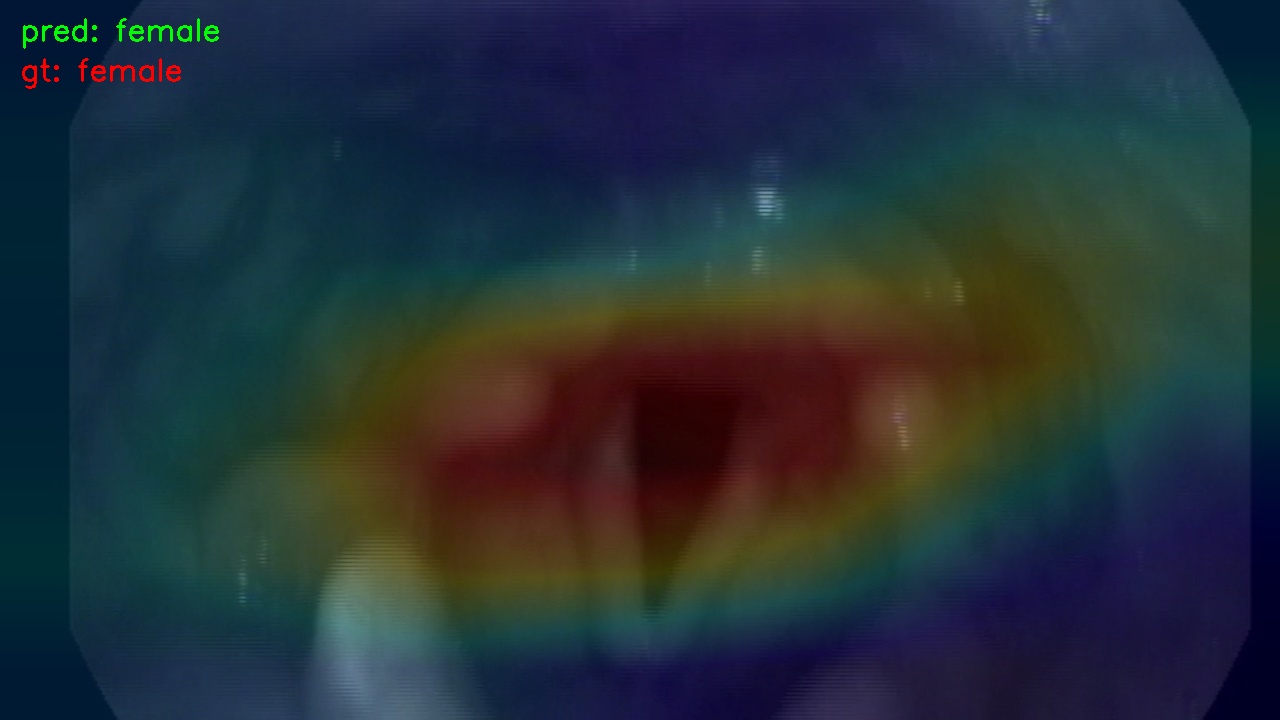}\\

\caption{The CAM visualization of gender prediction. ``pred'' stands for the predicted result and ``gt'' represents the ground truth. The left column demonstrates the maps for male patients and the right column illustrates the maps for female patients. The red color indicates the areas on the image have a high response for the predicted result and the blue color means the areas on the image have a low response for the predicted result.}
\label{fig:1}
\end{figure}

\begin{figure}[H]
\includegraphics[width=.48\textwidth]{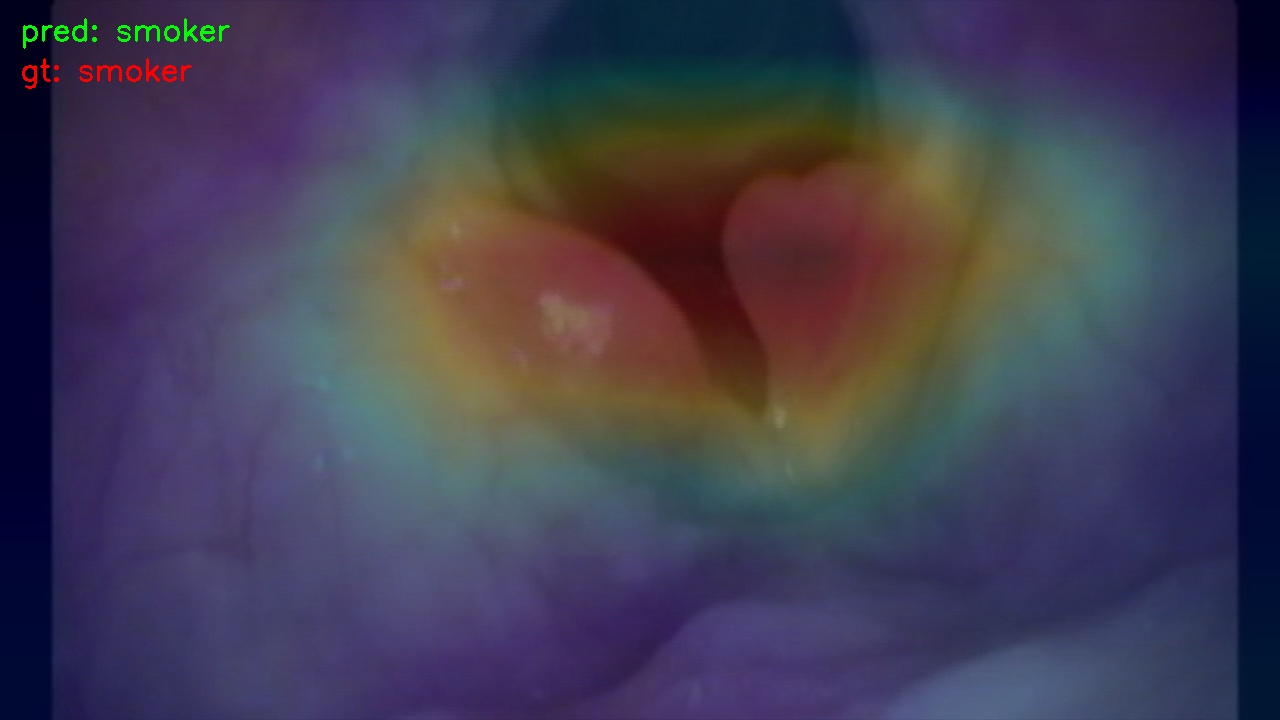}
\includegraphics[width=.48\textwidth]{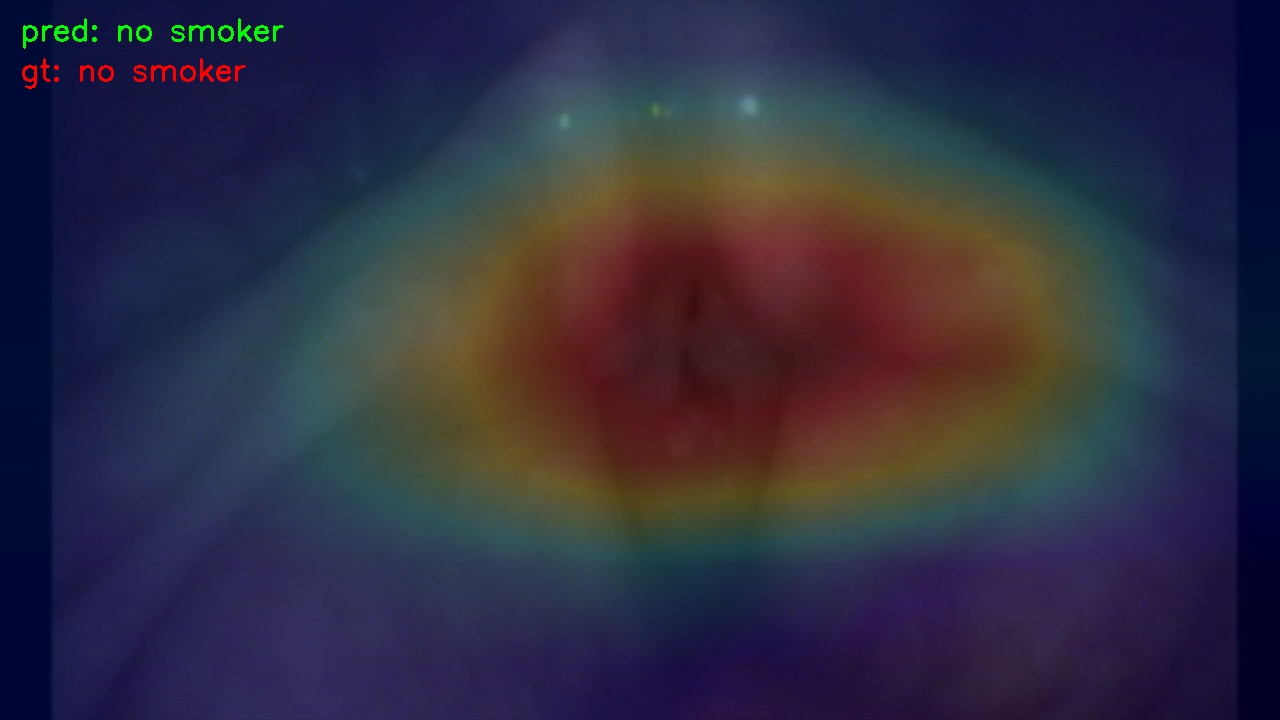}\\
\includegraphics[width=.48\textwidth]{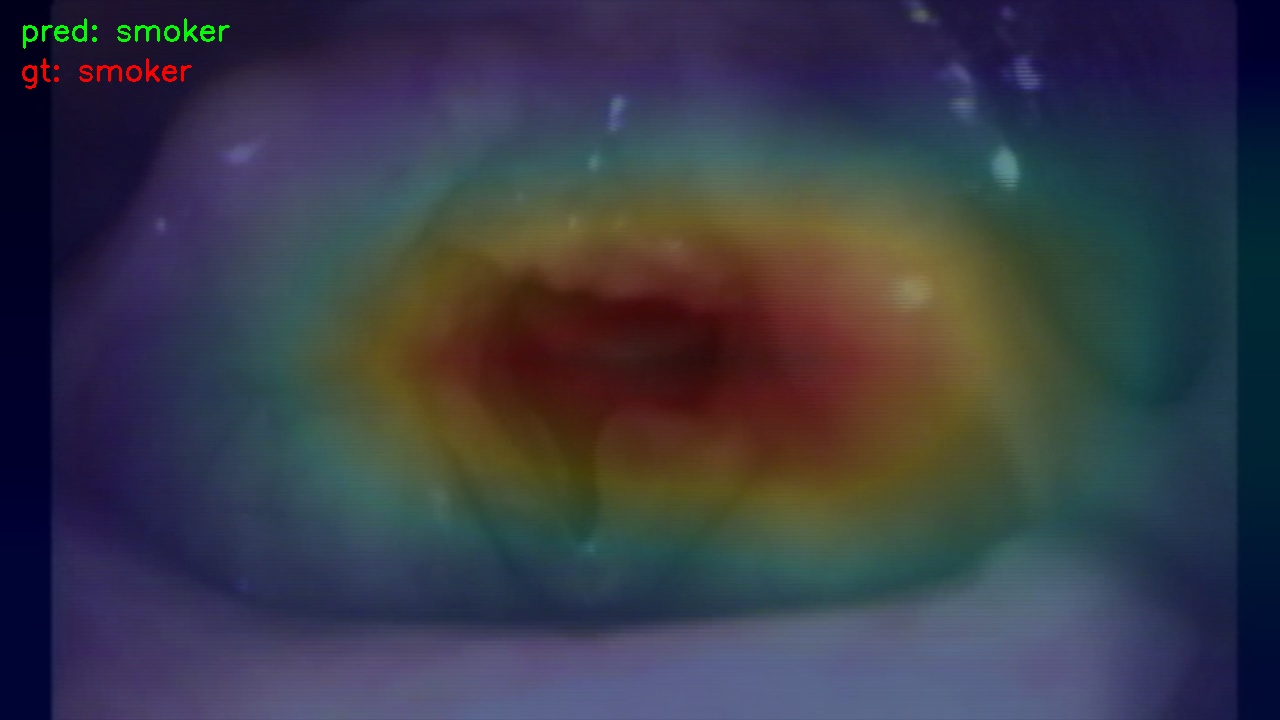}
\includegraphics[width=.48\textwidth]{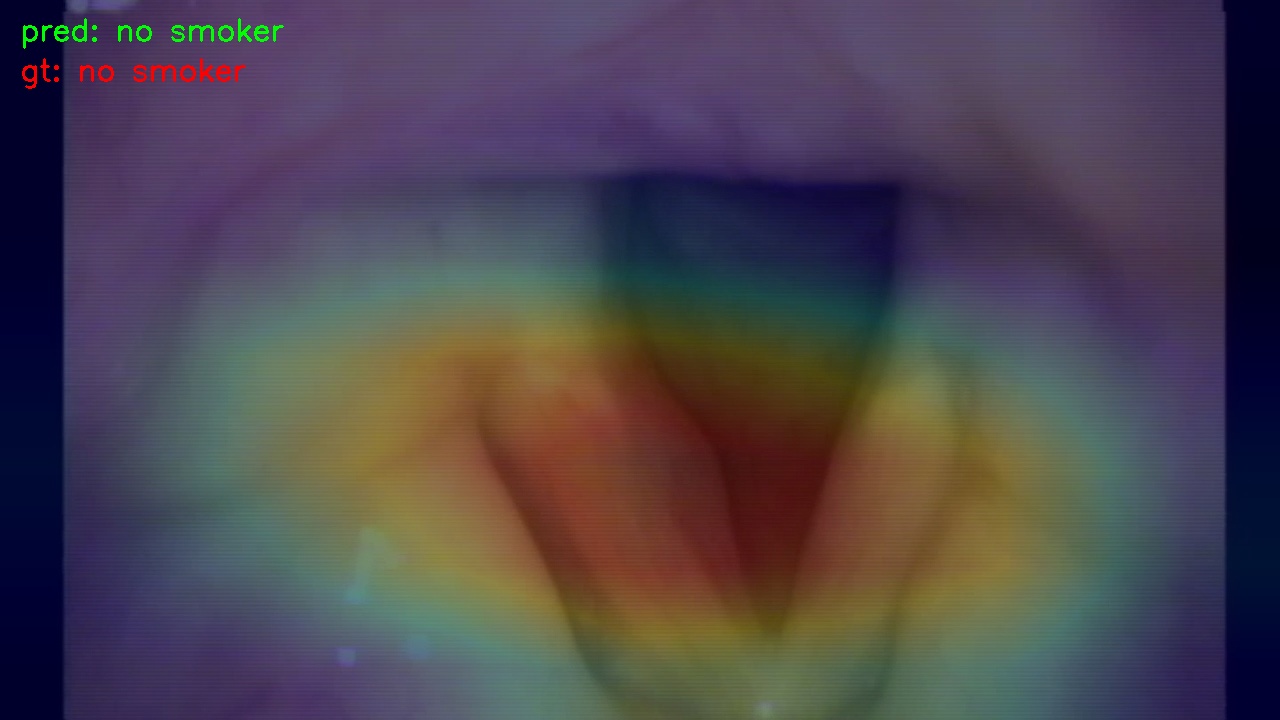}\\

\caption{The CAM visualization of smoking history prediction. ``pred'' stands for the predicted result and ``gt'' represents the ground truth. The left column demonstrates the maps for male patients and the right column illustrates the maps for female patients. The red color indicates the areas on the image have a high response for the predicted result and the blue color means the areas on the image have a low response for the predicted result.}
\label{fig:2}
\end{figure}

\begin{figure}[H]
\centering
\includegraphics[width=.48\textwidth]{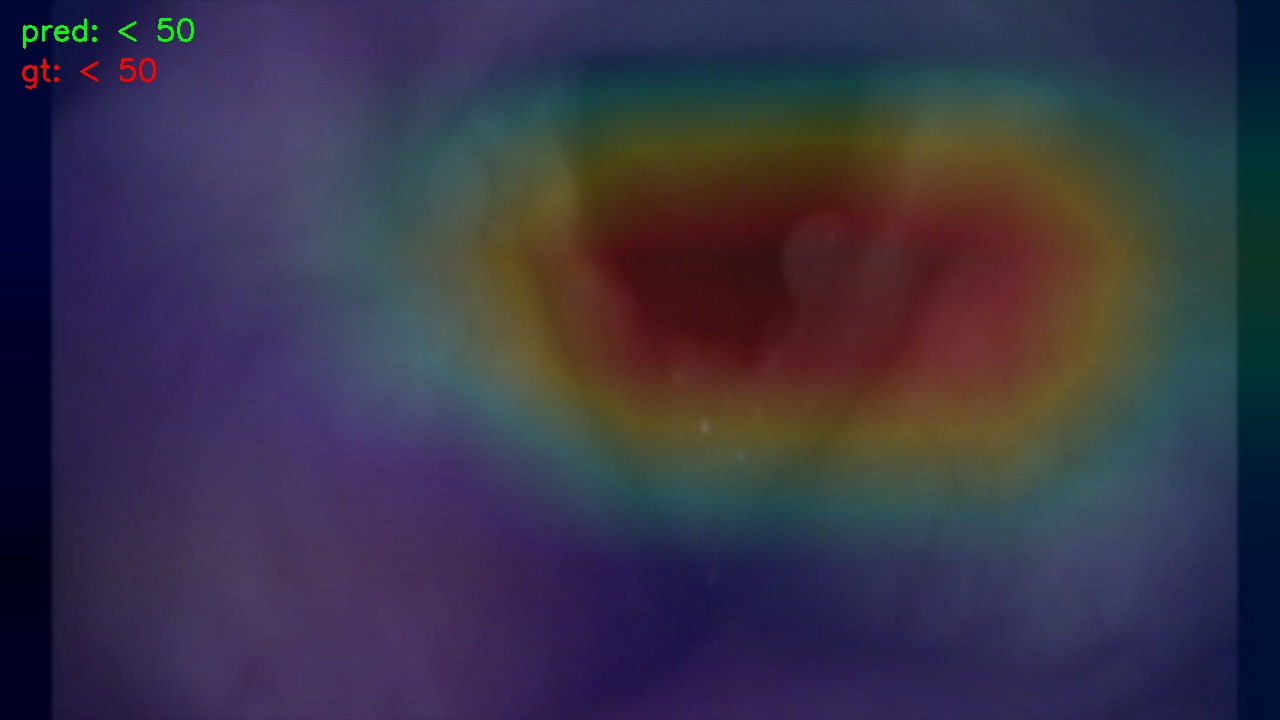}
\includegraphics[width=.48\textwidth]{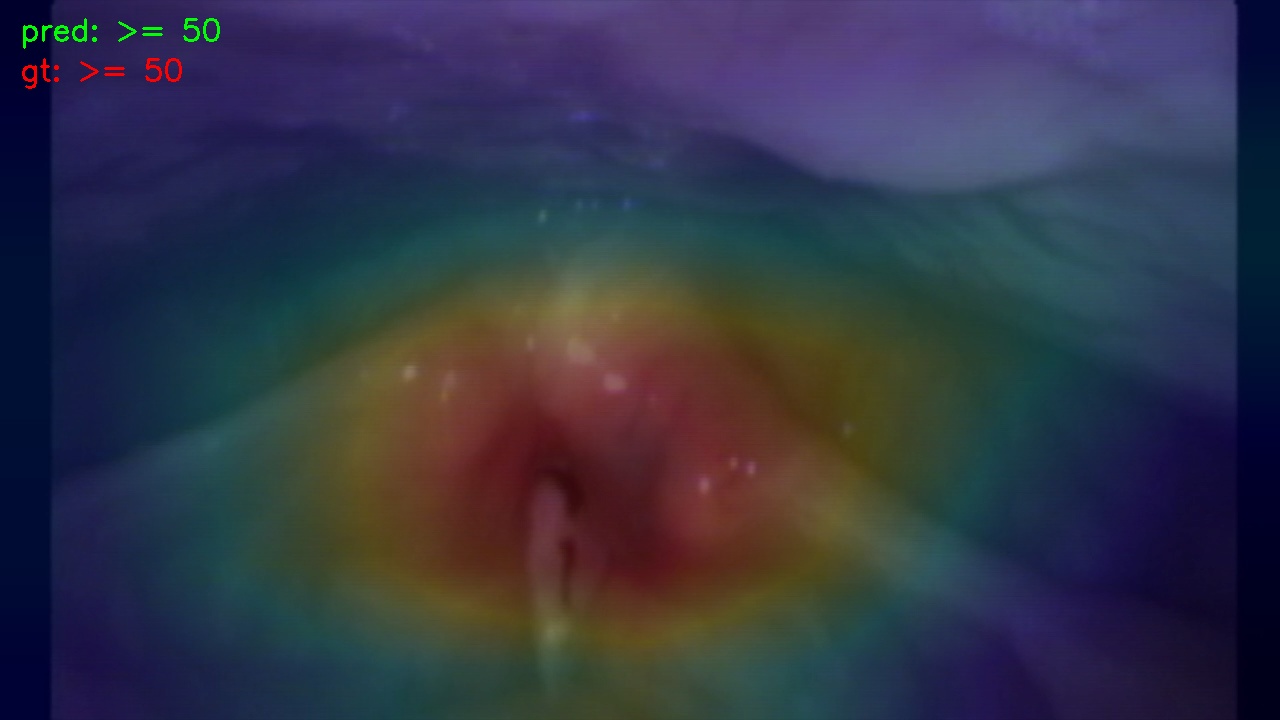}
\includegraphics[width=.48\textwidth]{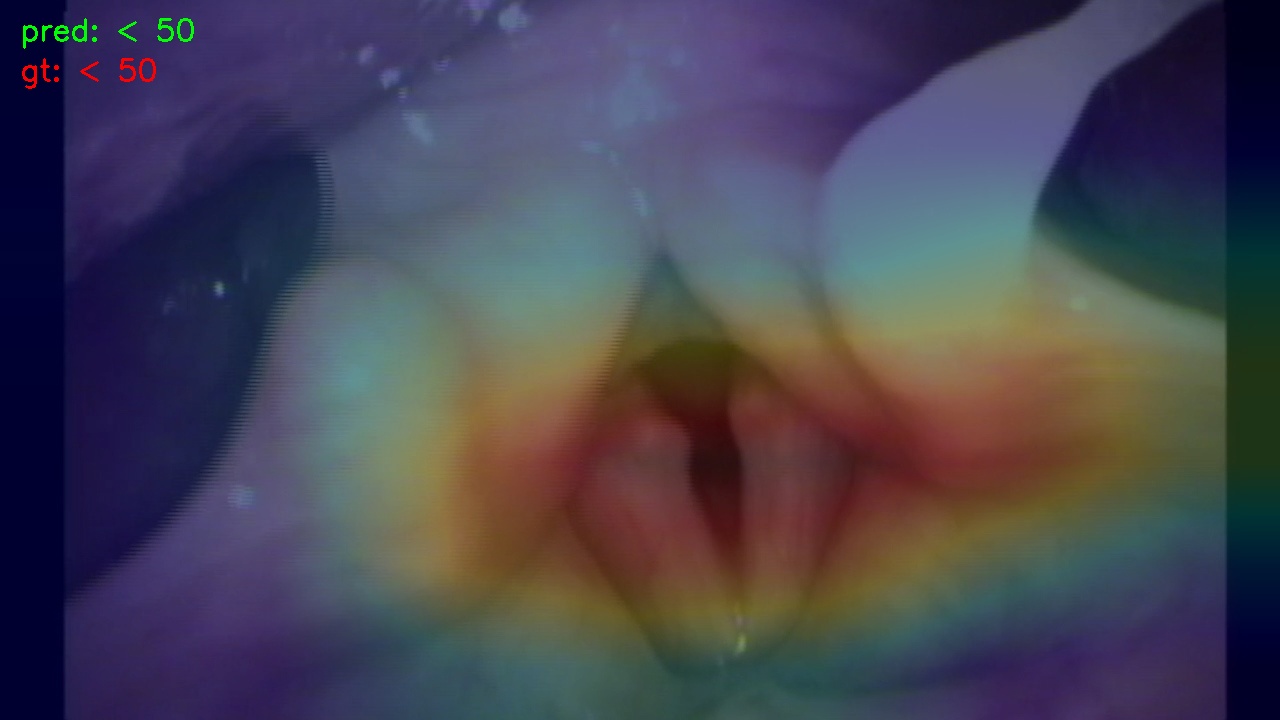}
\includegraphics[width=.48\textwidth]{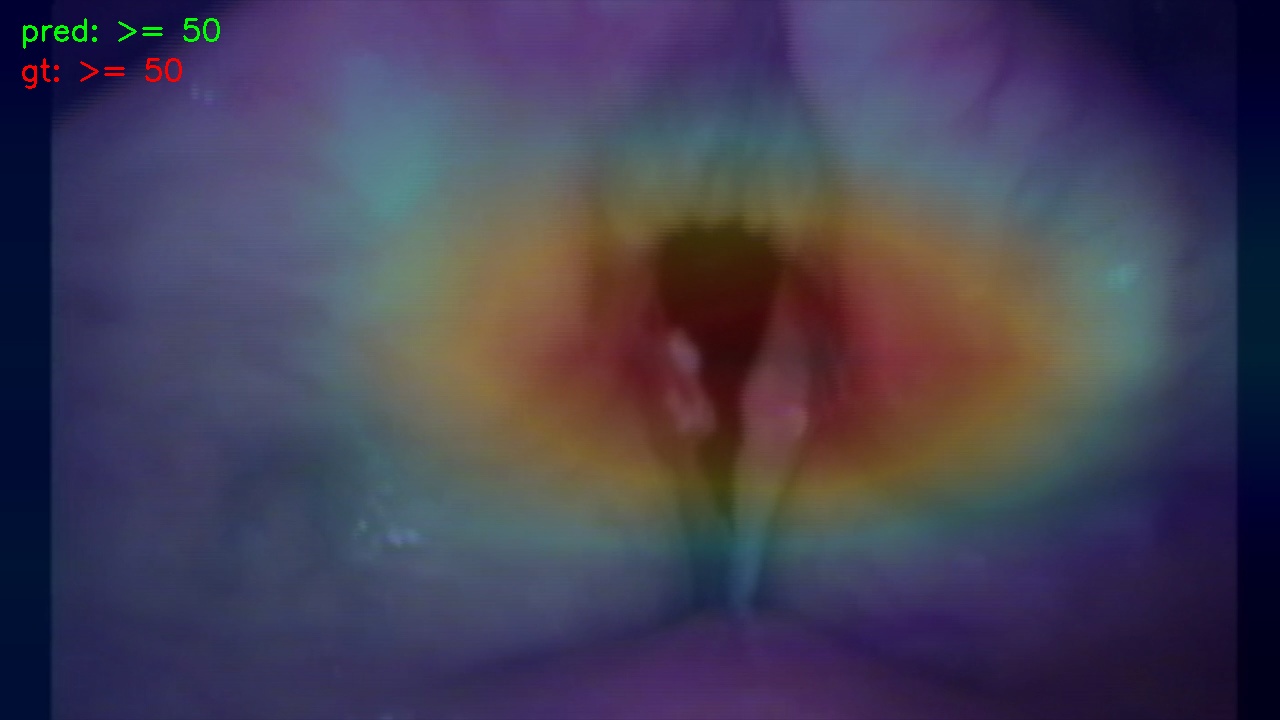}

\caption{The CAM visualization of age prediction. ``pred'' stands for the predicted result and ``gt'' represents the ground truth. The left column demonstrates the maps for male patients and the right column illustrates the maps for female patients. The red color indicates the areas on the image have high response for the predicted result and the blue color means the areas on the image have low response for the predicted result.}
\label{fig:3}
\end{figure}

In order to illustrate the response strength within different areas of images that correspond to the prediction results, a Classification Activation Map (CAM)~\cite{zhou2016learning} was extracted. Only results obtained by ResNet-50 were utilized for visualization, as similar results were obtained by other learning models. The CAMs for the prediction of gender, smoking history, and age are illustrated in Figure~\ref{fig:1}, Figure~\ref{fig:2} and Figure~\ref{fig:3}, respectively.

For visualization, CAM maps were overlayed on top of the original laryngeal images with a ratio of 2:5 so that the high-response areas on the original images could be easily recognized. The red color indicates high response and the blue color represents low response. High response areas correspond to areas that contribute more to the prediction~results.

Figure~\ref{fig:1} shows the CAM map for gender prediction. The left images are from male patients and the right images are from female patients. The high response areas were similar in both male and female images involving the true and false focal folds and partially the arytenoids. For smoking history and age prediction, the corresponding CAMs are illustrated in Figures~\ref{fig:2} and \ref{fig:3}, respectively. The high response areas had increased arytenoid involvement and were less focused on the vocal folds.

\section{Conclusions}

This is the first study to employ deep learning models with computer visualization to predict the gender, smoking history, and age of patients from laryngeal images. The deep learning models tested achieved consistent and promising results for these tasks. By visualizing the CAMs of the laryngeal images, the high response areas were focused primarily around the true and false vocal folds, which indicates that these areas may exhibit subtle differences among patients of different genders, ages, and smoking statuses. 

While we have annotated a laryngoscopic dataset in this study, the trained models may exhibit poor generalizability due to the relatively small scale of the dataset. To mitigate this limitation, it is essential to explore strategies that enhance service continuity in medical image classification. One viable solution is the design of self-organized systems~\cite{conti2020novel} that can dynamically optimize model parameters based on the scalability of the dataset, thereby improving the reliability and adaptability of the system. In our future studies, we will integrate the findings of this study into a comprehensive model for laryngeal disease classification by leveraging multi-modality learning techniques to effectively combine information from various sources, leading to more accurate and reliable diagnostic outcomes. We believe that these advancements will contribute to improved diagnostic capabilities and ultimately benefit patient care.




\vspace{6pt} 




\funding{This work was partly supported in part by the Natural Sciences and Engineering Research Council of Canada (NSERC) under grant no. ALLRP 576612-22, and the National Institutes
of Health (NIH) under grant no. 1R03CA253212-01.}

\dataavailability{ {content.}} 


\conflictsofinterest{The authors declare that they have no competing interests.}


\begin{adjustwidth}{-\extralength}{0cm}

\reftitle{References}





\PublishersNote{}
%


\end{adjustwidth}
\end{document}